\documentclass[letterpaper]{article} 
\usepackage{aaai2027}  
\usepackage[hyphens]{url}  
\usepackage{graphicx} 
\usepackage{natbib}  
\usepackage{caption} 
\usepackage{algorithm}
\usepackage{algorithmic}

\usepackage{newfloat}
\usepackage{listings}
\DeclareCaptionStyle{ruled}{labelfont=normalfont,labelsep=colon,strut=off} 
\floatstyle{ruled}
\newfloat{listing}{tb}{lst}{}
\floatname{listing}{Listing}

\usepackage{booktabs}

\usepackage{pifont}
\usepackage{amsmath}
\usepackage{amssymb}
\usepackage{multirow}

\nocopyright

\title{Task-Oriented Key-Layer KV Communication for Efficient Latent Multi-Agent Collaboration}
\author{
    Dongsen Zhang\textsuperscript{\rm 1},
    Peipei Li\textsuperscript{\rm 1}\corresponding,
    Zekun Li\textsuperscript{\rm 2},
    Wenjun Xu\textsuperscript{\rm 1}
}
\affiliations{
    \textsuperscript{\rm 1}Beijing University of Posts and Telecommunications, \textsuperscript{\rm 2}University of California, Santa Barbara\\
    \{dongsenzhang,lipeipei\}@bupt.edu.cn
}

\begin{document}

\maketitle

\begin{abstract}
Large language model-based multi-agent systems improve complex problem solving through collaboration, while latent communication directly transmits model internal states to avoid the high inference costs of natural language. However, existing KV-based latent communication methods prioritize sender-side state fidelity, leading to substantial communication and computation overhead and potentially introducing redundant information. To address these limitations, we revisit latent communication from a task-oriented perspective, shifting its objective from sender-side state fidelity to receiver-side task sufficiency. Under this formulation, we
propose KITE, a training-free framework for task-oriented key-layer KV communication. KITE identifies a task-effective key layer using a receiver trajectory distortion criterion, transmits only the latent working memory associated with the key layer, and further uses the same layer as the entry point for autoregressive latent reasoning. Experiments on seven benchmarks across two model families and three model scales show that, compared with full-layer KV communication, KITE reduces communication volume by 28-36$\times$, achieves up to 3$\times$ end-to-end inference speedup, and improves accuracy by up to 23.3 percentage points.

\end{abstract}


\section{Introduction}

Large language model (LLM)-based multi-agent systems (MAS) coordinate multiple agents through communication, enabling them to combine complementary capabilities and solve complex problems beyond the capacity of a single model \cite{coll_power}. This collaborative paradigm has been applied to diverse downstream tasks, including science reasoning \cite{science-reasoning1,science-reasoning2}, code generation \cite{coding1,coding2}, and embodied intelligence \cite{Embodied1,Embodied2}. Most LLM-based MAS rely on text-based communication, which requires agents to repeatedly convert internal reasoning into text and received text back into model-internal representations.

Recent work explores latent communication, in which agents directly exchange model-internal representations \cite{c2c,liu2025deliberation}. LatentMAS shares an evolving latent working memory across agents by transmitting KV cache from all Transformer \cite{transformer} layers that encode both the input context and newly generated latent thoughts \cite{latentmas}. To reduce communication cost, KVComm selects a subset of KV layers based on attention importance scores \cite{kvcomm}.

\begin{figure}[t]
    \centering
    \includegraphics[width=0.48\textwidth,trim=0 1600 3500 0,clip]{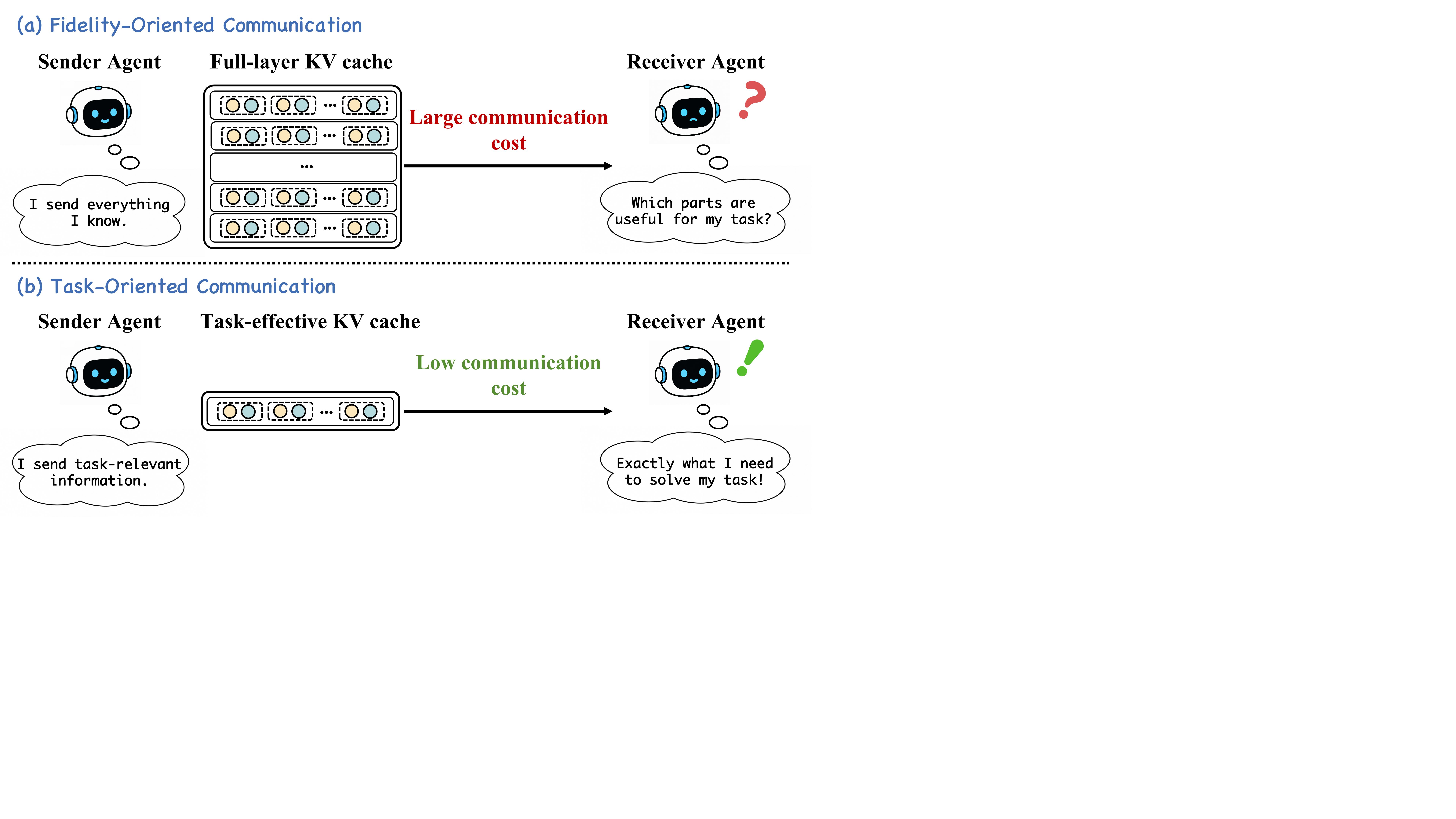}
    \caption{\textbf{(a) Fidelity-oriented communication} transmits all layer-wise KV cache to preserve sender-side state fidelity. \textbf{(b) Task-oriented communication} transmits only the task-effective KV cache to preserve receiver-side task utility.}
    \label{fig:0}
\end{figure}

Despite these advances, existing methods remain primarily focused on sender-side state fidelity, whereas the objective of multi-agent collaboration is receiver-side task completion. These objectives are not necessarily aligned. We therefore reformulate latent communication from a \textbf{sender-side state fidelity problem} to a \textbf{receiver-side task sufficiency problem}. Accordingly, the value of communicated latent states should be determined not by how faithfully they preserve sender-side states, but by their task utility to the receiver.

We study this problem through the lens of task-oriented communication \cite{task-orientedComm1,task-orientedComm2}, which aims to transmit task-relevant information for downstream task completion rather than recover all source data, as shown in Figure \ref{fig:0}. From this perspective, we argue that ideal latent communication should satisfy three properties: \textbf{\ding{172} Task Effectiveness}: the communicated states should preserve receiver-side task performance; \textbf{\ding{173} Communication Efficiency}: task performance should be preserved with minimal communication volume; and \textbf{\ding{174} Computational Efficiency}: the communicated states should be directly reusable by the receiver to avoid redundant computation.

Based on these principles, we propose \textbf{KITE}, a training-free framework for task-oriented key-layer KV communication. To evaluate receiver-side task sufficiency, KITE uses the receiver's reasoning trajectory under full-layer KV communication as a reference and measures the receiver trajectory distortion caused when communication is restricted to each individual layer's KV cache. Using this criterion, KITE identifies a model-specific key layer through one-time calibration on samples from multiple domains and uses it as the communication interface across downstream tasks. During collaboration, rather than using full-layer KV communication or attention-based KV-layer selection, agents exchange only the evolving latent working memory at this key layer. The same key layer also serves as the entry point for autoregressive latent reasoning, enabling the communicated states to be directly reused without recomputing the preceding Transformer layers.

To evaluate KITE, we conduct experiments on seven benchmarks spanning science reasoning, commonsense understanding, and code generation. Across two model families and three model scales, KITE consistently achieves higher accuracy than full-layer KV communication and attention-based KV-layer selection. Compared with full-layer KV communication, KITE reduces communication volume by 28$\times$ to 36$\times$, achieves up to a 3$\times$ end-to-end inference speedup, and improves accuracy by up to 23.3 percentage points. In summary, our work makes three key contributions:

\begin{itemize}
    
    \item \textbf{Task-oriented formulation and receiver-side criterion.} We introduce a task-oriented perspective for latent multi-agent communication, shifting the objective from sender-side state fidelity to receiver-side task sufficiency. Based on this formulation, we propose receiver trajectory distortion as a practical proxy for evaluating the task utility of layer-wise KV cache. This criterion identifies a model-specific key layer that generalizes across downstream tasks from diverse domains.
    
    \item \textbf{Unified key-layer communication and computation framework.} We propose KITE, a training-free framework that communicates only the latent working memory associated with the selected key layer and uses the same layer as the entry point for autoregressive latent reasoning. This unified design reduces both inter-agent communication overhead and redundant intra-agent computation.

    \item \textbf{Joint performance and efficiency gains.} Extensive experiments across diverse benchmarks, model families, and backbone scales show that KITE consistently improves downstream task performance while substantially reducing communication and computation costs. Further analysis suggests that key-layer communication may act as an implicit information filter.
    
\end{itemize}

\section{Related Work}

\subsection{LLM Communication in Latent Space}
Recent studies on multi-agent collaboration have shifted the communication medium from natural language text to model internal representations. CIPHER \cite{ciphers} communicates through the embedding space by transferring a weighted average of the vocabulary token embeddings from one LLM to another. Hidden states provide another latent communication channel: AC \cite{AC} injects the sender’s final token hidden state into the corresponding position of the receiver, while ThoughtComm \cite{ThoughtComm} employs an autoencoder to extract and route latent thoughts from hidden representations. Interlat \cite{interlat} directly communicates aligned sequences of model generated last layer hidden states between agents. KV caches have also been explored as a communication medium. C2C \cite{c2c} projects and fuses the sender’s contextual KV cache into the receiver, whereas LatentMAS \cite{latentmas} communicates layer-wise KV cache accumulated during both prompt prefill and latent reasoning, thereby forming an evolving shared latent working memory. Unlike these KV-based approaches that communicate information from all model layers, KITE transmits only the latent working memory associated with a single task-effective key layer, substantially reducing communication overhead for efficient latent multi-agent collaboration.

\subsection{KV Cache Optimization}
A substantial body of work improves KV cache efficiency through sharing and compression. Methods \cite{gqa,mqa} focusing on the model's head allow multiple query heads to share the same key and value head to reduce memory bandwidth requirements. At the layer level, methods \cite{KVreuse1,KVreuse2} reuse KV cache from shallower layers in deeper layers to accelerate single LLM inference. Several works \cite{mla,KVcompress} have explored compressing the KV cache to reduce memory usage. These methods primarily optimize computation and memory within a single model and are orthogonal to KITE; they can be combined with KITE for further efficiency gains.

Another line of work reuses KV cache associated with shared content across multiple queries \cite{KVreuse1,KVreuse2}. DroidSpeak \cite{droidspeak} selectively recomputes the KV cache of a subset of layers while directly reusing the remaining cache for overlapping content. KVComm-Online \cite{kvcomm-online} corrects cache offsets for repeated contexts under different prefixes using historical deviations. Unlike these methods that accelerate prefilling by reusing KV cache associated with explicit shared text, KITE communicates the single layer KV cache of an evolving latent working memory across agents.

KVComm \cite{kvcomm} is the most closely related work to KITE. Its sender prefills a context without autoregressive generation and selects KV layers according to their attention scores during receiver prefill. Our work goes beyond this related work by 1) enabling a different type of communication, where the sender performs latent reasoning, 2) formulating latent communication from a task-oriented perspective centered on receiver-side task utility, and 3) aligning the communication layer with the latent reasoning entry point to reduce intra-agent computation.

\subsection{Layer Skipping in Inference}
Layer skipping reduces LLM inference cost by avoiding computations that contribute marginally to the final prediction. LayerDrop \cite{layerdrop} introduces structured dropout during training, allowing different subsets of layers to be skipped during inference. ShortGPT \cite{shortgpt} estimates layer importance from input–output similarity and skips low-importance layers. LaCo \cite{laco} compresses model depth through a layer-collapse operation that absorbs the parameter differences of several consecutive layers into a preceding layer. Beyond static pruning, MindSkip \cite{MindSkip} and FlexiDepth \cite{flexidepth} each employ a lightweight router to adaptively skip layers based on the input. POLAR \cite{PoLar} represents Transformer layers as contiguous computational modules and predicts an input-specific execution program that can skip these modules. Our latent injection targets latent reasoning in multi-agent collaboration, skipping layers that are not directly involved in inter-agent communication.

\section{Preliminary and Notations}

\subsection{Sequential Multi-Agent Collaboration}
We consider a multi-agent system comprising four agents $\mathcal{S} = \left \{ A_1, A_2, A_3,A_4 \right \} $, where each agent is instantiated from the same LLM with $L$ layers. Given an input query $q$, the agents collaborate in a sequential chain order \cite{zhang2024chain}: $A_1\to A_2\to A_3\to A_4$. Specifically, the planner agent $A_1$ outlines an initial plan for solving $q$, the critic agent $A_2$ evaluates the initial plan and provides feedback, the refiner agent $A_3$ revises the plan accordingly, and the solver agent $A_4$ produces the final response with the plan. The intermediate agents communicate through KV cache, while only the solver decodes the final output into natural language.

\subsection{KV Cache as Latent Working Memory}
Unlike existing selective KV sharing methods that exchange information only on prefilled input context across models, our setting involves latent working memory that evolves throughout the collaboration. Each non-terminal agent $A_i \in \left \{ A_1,A_2,A_3 \right \}$  performs $m$ latent reasoning steps, where reasoning unfolds through autoregressive generation of last-layer hidden representations without explicit decoding. The internal states produced during both prefill and latent reasoning are accumulated in the KV cache, forming a growing latent working memory. We denote the memory passed by agent $A_i$ as

$$
\mathcal{M}_i=\mathcal{M}_{i-1} \oplus \mathcal{M}_{i,\text{prefill}}\oplus \mathcal{M}_{i,\text{reasoning}},
$$

\noindent where $\mathcal{M}_{i-1} $ is the memory inherited from preceding agents, and $\mathcal{M}_{0}= \varnothing$ for the planner agent. $ \mathcal{M}_{i,\text{prefill}}$ and $\mathcal{M}_{i,\text{reasoning}}$ are the KV cache generated during prefill and latent reasoning, respectively, and $\oplus$ denotes concatenation along the sequence dimension. Agents communicate by forwarding this accumulated latent working memory to the next agent.

\section{Method}

\begin{figure*}[t]
    \centering
    \includegraphics[width=\textwidth,trim=0 00 00 0,clip]{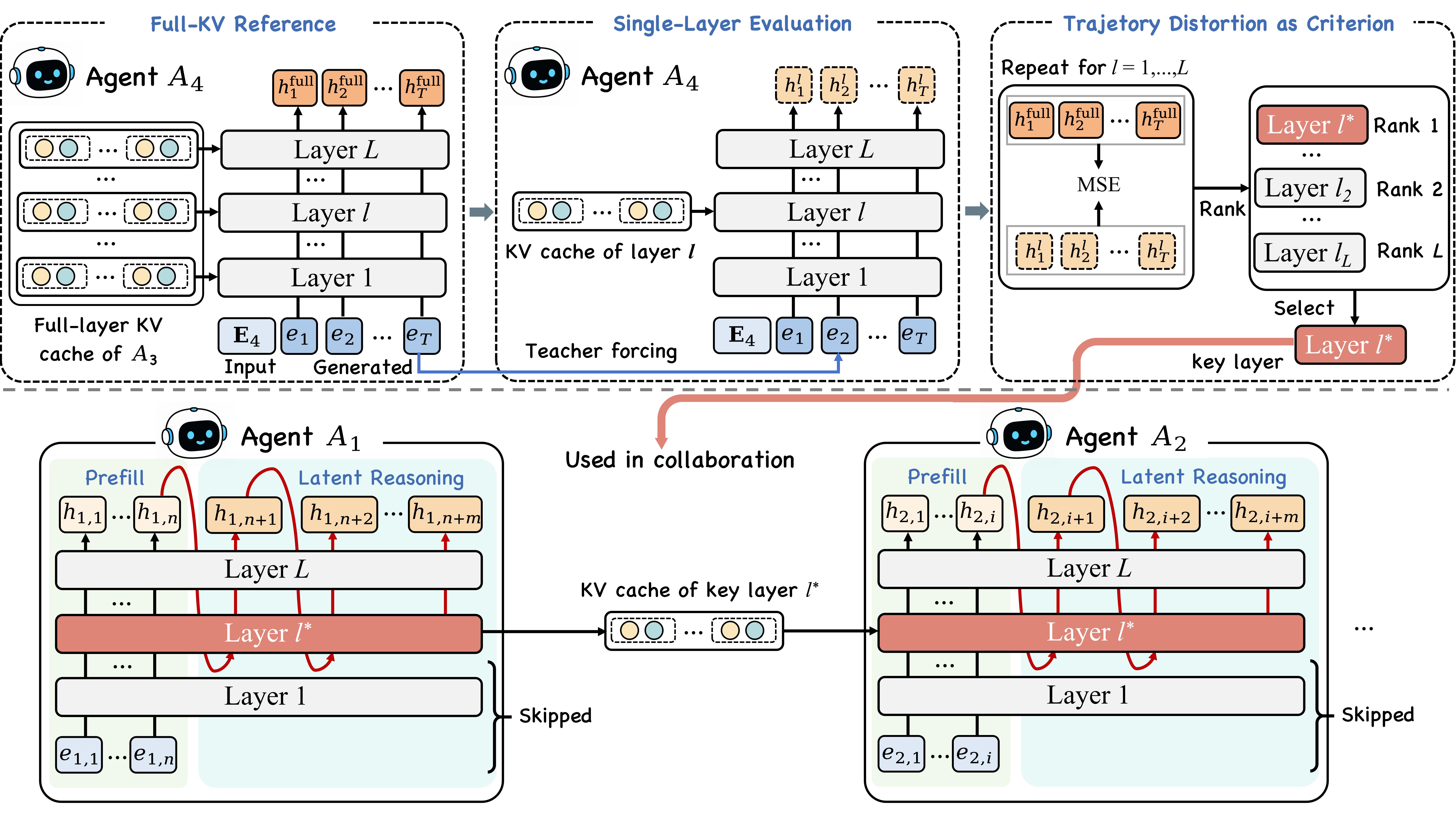}
    \caption{Overview of the KITE framework. \textbf{Key layer calibration (top)} identifies the task-effective KV layer by receiver trajectory distortion. \textbf{Agents communicate (bottom)} through the key layer latent working memory and use the same layer as the entry point for latent reasoning.}
    \label{fig:1}
\end{figure*}

We propose KITE, an efficient communication framework for latent collaboration that localizes both inter-agent communication and intra-agent computation at a task-effective key layer within the LLM. The framework is illustrated in Figure \ref{fig:1}. KITE is designed around three objectives: \ding{172} ensuring task effectiveness of information by preserving the receiver-side behavior, \ding{173} improving communication efficiency by transmitting only the KV cache of the selected key layer, and \ding{174} improving computation efficiency by skipping unnecessary forward computation in preceding layers.

\subsection{Key-layer KV Selection and Communication}

Task-oriented communication aims to transmit only the information required by the receiver to accomplish its task. This necessitates selecting task-effective  KV cache, as the task utility of KV is not uniform across LLM layers. Formally, given the full latent working memory from the preceding agent $\mathcal{M}_3 = \left \{(\mathbf{k}_3^l, \mathbf{v}_3^l)\right \}_{l=1}^L$ and input embeddings $\mathbf{E}_4$ containing the query $q$ and the instruction prompt, the objective of task-oriented communication is to select the minimal layer subset $S \subseteq \left \{ 1,...,L\right \}$ that enables the receiver to achieve the maximum task utility. This can be formulated as a bi-objective optimization problem as


\begin{equation}
\operatorname*{arg\,max}_{S\neq\emptyset}
\biggl(
U\!\left(
A_4\!\left(\mathbf{E}_4,\mathcal{M}_3^{S}\right)
\right),
-|S|
\biggr),
\end{equation}

\noindent where $U(\cdot)$ is a task completion metric (e.g., accuracy), and $A_4(\mathbf{E}_4,\mathcal{M}_3^{S})$ denotes the output of the solver agent given the input embeddings $\mathbf{E}_4$ and the selected KV cache $\mathcal{M}_3^{S}=\left \{(\mathbf{k}_3^l, \mathbf{v}_3^l)\right \}_{l\in S}$. The first objective promotes receiver-side task effectiveness, while the second minimizes the communication volume. However, directly solving this objective is intractable because evaluating task utility may require task labels or executable feedback, and exhaustively evaluating all layer subsets incurs combinatorial cost. This raises a fundamental question: how to identify the task effectiveness of KV cache?

We use receiver trajectory distortion as a label-free proxy for estimating the task utility of layer-wise KV caches. The receiver trajectory is represented by the sequence of last hidden states. These hidden states accumulate contextual and reasoning information, and provide a compact representation of the model behavior \cite{hiddenstate1,hiddenstate2}. Intuitively, if a layer’s KV cache produce a receiver trajectory close to that obtained with full KV cache, it should encode information critical to task completion. Last hidden states deviation thus serves as a practical proxy for the communication value of a KV subset. This is empirically validated in Section ~\ref{sec:Selection_Strategy_Effectiveness}.

Specifically, we use the solver agent’s last hidden states during autoregressive generation with full KV as the reference, denoted as $\mathbf{H}_4^{\text{full}} = \left \{ \mathbf{h}_{4,t}^{\text{full}} \right \}_{t=1}^T$. For each layer $l$, we transmit only its KV cache and rerun the solver with teacher forcing: at each decoding step, the solver is fed the same token generated with full KV, rather than its own predicted token. This yields an aligned single KV layer trajectory $\mathbf{H}_4^{l} = \left \{ \mathbf{h}_{4,t}^{l}  \right \}_{t=1}^T$. We then measure the trajectory distortion using mean squared error:

\begin{equation}
    D_l = \frac{1}{Td}\sum_{t=1}^{T}  \left \| \mathbf{h}_{4,t}^{\text{full}} -\mathbf{h}_{4,t}^{l} \right \| _2^2,
\end{equation}

\noindent where $T$ is the number of autoregressive decoding steps and $d$ is the hidden state dimension. A lower trajectory distortion indicates that the layer better preserves the receiver behavior.


LLM layers exhibit a hierarchical functional organization \cite{hierarchical1,hierarchical2}, suggesting that effective KV layers should remain consistent across domains. Therefore, we construct a calibration set using datasets from two distinct domains to identify KV layers that are task-effective across domains. We rank all layers by their average receiver trajectory distortion over the calibration set and observe that communicating only the KV cache of the highest-ranked layer is sufficient for the receiver to complete the calibration tasks. Accordingly, we set the communicated layer subset size to $|S|=1$. For each model, we select the highest-ranked layer on the calibration set as the model-specific key layer $l^*$ and reuse it for all downstream tasks.


Each sender agent $A_i$ transmits only the latent working memory $\mathcal{M}_i^{l^*} = (\mathbf{k}_i^{l^*}, \mathbf{v}_i^{l^*})$ at the selected key layer $l^*$. Consequently, the entire collaboration proceeds through a single key layer communication path.

\subsection{Layer Aligned Latent Injection}

We introduce key layer aligned latent reasoning within the agent. In multi-agent latent collaboration, communication does not rely on natural language, allowing each non-terminal agent to perform autoregressive reasoning directly through last hidden states without decoding them into explicit tokens. Prior analyses \cite{PoLar,flexidepth} suggest that pretrained LLMs can support valid latent computations along non-standard execution paths that deviate from the conventional fixed-depth forward pass. Since information from non-key layers is not communicated, we skip the preceding layers and inject the last hidden state directly into the selected key layer as input. The key layer serves as the entry point for latent reasoning, with forward propagation performed only from the key layer to the final layer.

Specifically, given the key layer latent working memory $\mathcal{M}_{i-1}^{l^{*}}$ and $n$ input embeddings $\mathbf{E}_i =\left \{ \mathbf{e}_{i,j} \right \}_{j=1}^n$, each non-terminal agent $A_i$ performs standard prefill through all $L$ layers to obtain the last hidden state sequence $\left \{ \mathbf{h}_{i,j}  \right \}_{j=1}^n$. During latent reasoning, we use the final states $\mathbf{h}_{i,n} $ as the input to the key layer, and compute the next last hidden states from  $l^*$ onward:

\begin{equation}
    \mathbf{h}_{i,n+1} = F_{l^{*}:L}(\mathbf{h}_{i,n},\mathcal{M}_{i-1}^{l^{*}},\mathcal{M}_{i,\text{prefill}}),
\end{equation}

\noindent where $F_{l^{*}:L}$ denotes the forward computation from the key layer $l^*$ to the last layer. We repeat this process autoregressively for $m$ steps, yielding the key layer latent reasoning KV cache $\mathcal{M}_{i,\text{reasoning}}^{l^{*}}$.

\subsection{Complexity Analysis}
\label{Complexity Analysis}

\textbf{Communication Volume.} KITE transmits KV cache from only one of the $L$ layers, reducing the communication volume to $\frac{1}{L}$of that required by LatentMAS.

\noindent \textbf{Computational Complexity.} Let $c$ denote the length of the incoming latent working memory, $d$ the hidden dimension, and $r$ the number of tokens generated by $A_4$. In the prefill stage, the computational savings of KITE over LatentMAS for each agent is on the order of $O((L-1)cnd)$. In the generation stage, the computation margin of KITE over LatentMAS for $A_i$ and $A_4$ is on the order of $O((l^*-1)(md^2+nmd+m^2d)+(L-1)cmd)$ and $O((L-1)crd)$, respectively.

\section{Experiments}

\begin{table*}[t]
\centering
\resizebox{\textwidth}{!}{%
\begin{tabular}{lc|ccccc|ccccc|ccccc}
\toprule[1.2pt]
\multirow{2}{*}{\textbf{Tasks}}      & \multirow{2}{*}{\textbf{Metrics}} & \multicolumn{5}{c|}{\textbf{Llama-3.2-3B-Instruct}} & \multicolumn{5}{c|}{\textbf{Qwen3-4B}}                              & \multicolumn{5}{c}{\textbf{Qwen3-8B}}                             \\ \cmidrule(lr){3-17} 
                                     &                                   & TMAS  & LMAS & KVC & KITE  & KITE--LI & TMAS & LMAS      & KVC & KITE           & KITE--LI   & TMAS & LMAS      & KVC & KITE          & KITE--LI   \\ \midrule
\multirow{4}{*}{\textbf{ARC-E}}      & Acc.~$\uparrow$                              & 75.7     & 75.2      & 74.9   & 77.8  & \textbf{79.1}        & 97.6    & 95.9           & 88.6   & \textbf{97.7}  & 97.3          & 98.3    & 97.5           & 87.0   & \textbf{98.4} & 97.9          \\
                                     & Volume ~$\downarrow$                            & 4.6      & 85.6      & 42.8   & \textbf{3.0}   & 3.0         & 3.9     & 100.9          & 50.4   & \textbf{2.8}   & 2.8           & 7.1     & 100.9          & 50.4   & \textbf{2.8}  & 2.8           \\
                                     & Time ~$\downarrow$                              & 10.0     & 1.2       & 1.2    & \textbf{0.8}   & 0.9         & 18.7    & 4.1            & 7.7    & \textbf{3.4}   & 4.0           & 21.9    & \textbf{4.6}   & 6.4    & 5.2           & 5.6           \\
                                     & Token                              & 1930     & 239       & 217    & 234   & 227         & 1840    & 271            & 497    & 376   & 379           & 2182    & 324   & 427    & 401           & 404          \\
                                     \midrule
\multirow{4}{*}{\textbf{ARC-C}}      & Acc.~$\uparrow$                              & 61.0     & 60.4      & 58.5   & \textbf{67.8}  & 67.3        & 94.2    & 91.2           & 80.2   & 92.8  & \textbf{93.0}          & 95.1    & 93.7           & 80.0   & \textbf{94.2} & 94.2          \\
                                     & Volume ~$\downarrow$                            & 4.6      & 91.5      & 45.7   & \textbf{3.2}   & 3.2         & 4.2     & 108.7          & 54.1   & \textbf{3.0}   & 3.0           & 7.6     & 108.7          & 54.3   & \textbf{3.0}  & 3.0           \\
                                     & Time ~$\downarrow$                              & 10.6     & 1.5       & 1.2    & \textbf{0.9}   & 1.1         & 22.7    & 6.0            & 8.9    & \textbf{5.6}   & 5.7           & 27.3    & \textbf{6.6}   & 8.3    & 7.2           & 7.5           \\
                                     & Token                              & 1965     & 252       & 239    & 248   & 246         & 1984    & 340            & 639    & 507   & 506           & 2335    & 386   & 483    & 513           & 511           \\ \midrule
\multirow{4}{*}{\textbf{HumanEval+}} & Acc.~$\uparrow$                              & 35.9     & 25.0      & 25.0   & \textbf{35.3}  & 33.5        & 89.0    & 76.8           & 40.8   & \textbf{86.5}  & 86.5          & 89.6    & 80.5           & 26.8   & \textbf{88.4} & 86.5          \\
                                     & Volume ~$\downarrow$                            & 6.4      & 132.7     & 66.3   & \textbf{4.7}   & 4.7         & 8.7     & 162.8          & 81.4   & \textbf{4.5}   & 4.5           & 13.4    & 162.8          & 81.4   & \textbf{4.5}  & 4.5           \\
                                     & Time ~$\downarrow$                              & 41.5     & 4.0       & 3.8    & \textbf{3.2}   & 3.3         & 132.4   & 68.9           & 119.6  & 58.4           & \textbf{57.6} & 111.2   & 62.6           & 58.5   & \textbf{57.4} & 62.3          \\
                                     & Token                              & 2392     & 407       & 425    & 325   & 310         & 4358    & 2244            & 4826    & 2444   & 2314           & 4125    & 2436   & 1410    & 2381           & 2491           \\ \midrule
\multirow{4}{*}{\textbf{MBPP+}}      & Acc.~$\uparrow$                              & 26.1     & 21.9      & 30.6   & \textbf{35.7}  & 34.6        & 78.8    & 61.6           & 13.7   & \textbf{76.9}  & 76.1          & 79.3    & 74.6           & 20.8   & 78.5 & \textbf{80.1}          \\
                                     & Volume ~$\downarrow$                            & 6.3      & 147.2     & 73.6   & \textbf{5.2}   & 5.2         & 7.5     & 188.2          & 94.1   & \textbf{5.2}   & 5.2           & 13.0    & 188.2          & 94.1   & \textbf{5.2}  & 5.2           \\
                                     & Time ~$\downarrow$                              & 39.9     & \textbf{3.81}      & 3.92   & 4.6  & 4.6         & 114.0   & 82.9           & 136.8  & \textbf{62.2}  & 62.7          & 132.0   & 76.7           & 100.0  & \textbf{61.7} & 63.4          \\
                                     & Token                              & 2314     & 375       & 404    & 437   & 327         & 3706    & 2374            & 5607    & 1975   & 2002           & 3904    & 1925   & 1971    & 2096           & 2144           \\ \midrule
\multirow{4}{*}{\textbf{GPQA}}       & Acc.~$\uparrow$                              & 19.1     & 21.2      & 19.1   & \textbf{22.7}  & 20.7        & 51.0    & 36.8           & 15.1   & \textbf{48.9}  & 44.9          & 57.0    & 46.4           & 23.7   & \textbf{49.4} & 47.9          \\
                                     & Volume ~$\downarrow$                            & 7.1      & 210.3     & 105.1  & \textbf{7.5}   & 7.5         & 15.7    & 262.3          & 131.1  & \textbf{7.2}   & 7.2           & 25.6    & 262.3          & 131.1  & \textbf{7.2}  & 7.2           \\
                                     & Time ~$\downarrow$                              & 69.2     & 52.3      & 50.6   & \textbf{45.1}  & 51.4        & 220.4   & 127.3          & 125.1  & \textbf{100.3} & 101.6         & 246.4   & \textbf{88.2}  & 95.4   & 114.6         & 114.6         \\
                                     & Token                              & 3320     & 1104       & 1041    & 1649   & 1681         & 6881    & 3602            & 4894    & 4609   & 4734           & 7255    & 2520   & 2461    & 4878           & 4741           \\ \midrule
\multirow{4}{*}{\textbf{AIME24}}     & Acc.~$\uparrow$                              & 3.3      & 3.3       & 6.6    & \textbf{13.3}  & 3.3         & 70.0    & 50.0           & 3.3    & \textbf{73.3}  & 60.0          & 76.6    & 46.6           & 26.6   & \textbf{63.3} & 63.3          \\
                                     & Volume ~$\downarrow$                            & 9.6     & 108.0     & 54.0   & \textbf{3.8}   & 3.8         & 26.4    & 131.1          & 65.5   & \textbf{3.6}   & 3.6           & 43.8    & 131.1          & 65.5   & \textbf{3.6}  & 3.6           \\
                                     & Time ~$\downarrow$                              & 156.9    & 488.1          &  713.6      & \textbf{162.6} & 227.3       & 523.6   & \textbf{386.2} & 810.8  & 570.4          & 622.2         & 541.6   & \textbf{422.8} & 637.4  & 600.8         & 607.8         \\ 
                                     & Token                              & 6314     & 11803       & 10819    & 4646   & 6586         & 12784    & 8333            & 15823    & 11535   & 12461           & 13151    & 8293   & 10988    & 11906           & 11986           \\ \midrule
\multirow{4}{*}{\textbf{AIME25}}     & Acc.~$\uparrow$                              & 0.0      & 0.0       & 0.0    & \textbf{3.3}   & 0.0         & 56.6    & 40.0           & 16.6   & \textbf{60.0}  & 50.0          & 60.0    & 36.6           & 26.6   & 53.3          & \textbf{56.6} \\
                                     & Volume ~$\downarrow$                            & 10.9     & 114.5     & 57.2   & \textbf{4.0}   & 4.0         & 33.2    & 154.1          & 77.0   & \textbf{4.2}   & 4.2           & 50.2    & 154.1          & 77.0   & \textbf{4.2}  & 4.2           \\
                                     & Time ~$\downarrow$                              & 169.3    &  419.2         & 728.9       & \textbf{263.9} & 301.7       & 631.7   & \textbf{450.1} & 640.8  & 709.5          & 697.7         & 594.0   & \textbf{512.6} & 633.7  & 758.0         & 758.6         \\
                                     & Token                              & 6266     & 10525       & 11733    & 7835   & 9037         & 15342    & 9714            & 11521    & 13877   & 13498           & 14509    & 11178   & 11556    & 14107           & 14158           \\ \bottomrule[1.2pt]
\end{tabular}%
}
\caption{Main results of KITE on 7 tasks. We report task accuracy (\%, \textbf{``Acc.''}), average communication volume (million elements, \textbf{``Volume''}), average inference time (seconds, \textbf{``Time''}), and average output token usage (\textbf{``Token''}). KITE–LI denotes the variant without latent injection, whose autoregressive latent reasoning follows LMAS. We \textbf{bold} the better performance over latent communication methods.}
\label{tab:main-result}
\end{table*}

\subsection{Experimental Setup}

\textbf{Datasets.} The calibration set consists of 300 samples from the medical reasoning dataset MedQA \cite{medqa} and 50 samples from the mathematical reasoning dataset GSM8K \cite{gsm8k}. Following LatentMAS, we comprehensively evaluate KITE on seven benchmarks covering both general-purpose and reasoning-intensive tasks: (i) Commonsense Reasoning, including ARC-Easy and ARC-Challenge \cite{arc}; (ii) Math \& Science Reasoning, including GPQA-Diamond \cite{gpqa}, AIME24 \cite{maxwelljia2024aime}, and AIME25 \cite{mathai2025aime}; and (iii) Code Generation, including HumanEval-Plus and MBPP-Plus \cite{mbppANDhumaneval}.

\noindent \textbf{Models and Baselines.}  We evaluate KITE on two LLM families at different model scales: Qwen3-4B, Qwen3-8B \cite{yang2025qwen3}, and Llama-3.2-3B-Instruct \cite{grattafiori2024llama}. We compare KITE against three baselines: (i) Text-based MAS (TMAS), where agents exchange information through natural language; (ii) LatentMAS (LMAS) \cite{latentmas}, which communicates KV cache from all layers, and uses aligned last hidden states as input embeddings, performing each latent reasoning step from the first layer; (iii) KVComm (KVC) \cite{kvcomm}, which communicates KV cache from half of the model layers, with the selected layers determined separately for each benchmark based on attention score.

\noindent \textbf{Metrics and Implementation Details.} We evaluate task sufficiency, communication efficiency, and computation efficiency using accuracy, average communication volume per sample, and average inference time per sample, respectively. For TextMAS, communication volume is measured as the total number of scalar elements in the transmitted token embeddings. For latent communication methods, it is measured as the total number of scalar elements in the transmitted KV cache. We also report average output token usage per sample. Following LMAS \cite{latentmas}, we adjust the latent reasoning steps and the maximum output length for each task according to its relative difficulty. We set the latent reasoning steps to 20 for ARC-Easy, ARC-Challenge and HumanEval+, and 80 for GPQA, AIME24/25 and MBPP+. We set the maximum length to 2,048 tokens for ARC-Easy and ARC-Challenge, 8,192 tokens for GPQA, HumanEval+ and MBPP+, and 20,000 tokens for AIME24/25.

\subsection{Main Results}

Table \ref{tab:main-result} reports the performance of KITE on seven benchmarks across three models of different scales. We make three main observations: (i) KITE consistently outperforms latent communication baselines in task accuracy across model–benchmark combinations, with an improvement of up to 23.3 percentage points over the stronger latent baseline. KITE can achieve comparable performance to TMAS, and outperform it on several tasks. These results suggest that key-layer KV communication may have an implicit regularization effect, helping the receiver focus on task-effective information. We further analyze this effect in Section \ref{accuracy improvement analysis}. (ii) KITE substantially reduces communication volume. Consistent with the analysis in Section \ref{Complexity Analysis}, its communication volume advantage scales linearly with model depth. Compared with LMAS, KITE reduces communication volume by 28$\times$ on Llama-3.2-3B-Instruct and 36$\times$ on Qwen3-4B/8B. Compared with KVC, the corresponding reductions are 14$\times$ and 18$\times$, respectively. (iii) KITE generally improves inference efficiency, achieving up to a 3$\times$ speedup over LMAS. On some reasoning-intensive tasks such as GPQA and AIME24/25, KITE generates more output tokens and consequently requires longer inference time, while achieving substantially higher accuracy. This suggests that the task-relevant information communicated by KITE enables the solver to sustain more complete reasoning before producing the final answer. Moreover, KITE is faster than KITE-LI in 17 of the 21 settings, demonstrating the efficiency benefit of layer-aligned latent injection.

\subsection{Selection Strategy Effectiveness}

To evaluate the effectiveness of our selection criterion, we randomly sample 10 layers for each model and compute the Pearson correlation coefficients between their trajectory-distortion scores and performance under single-layer KV communication across four tasks from diverse domains. We compare against the attention-based criterion in KVC. The results are shown in Table \ref{tab:correlation}. Trajectory distortion exhibits strong correlations with task accuracy across all model–task combinations. The advantage of trajectory distortion criterion is pronounced on Qwen3-4B and Qwen3-8B, where the average correlation coefficients reach -0.780 and -0.855. Notably, the correlations obtained by trajectory distortion remain strong on both in-domain and out-of-domain tasks. These results suggest that receiver trajectory distortion provides a practical proxy for estimating the task utility of layer-wise KV caches.

\begin{table}[t]
\centering
\resizebox{0.48\textwidth}{!}{%
\begin{tabular}{lcccc}
\toprule[1.2pt]
\textbf{Tasks}                       & \textbf{Methods} & \textbf{Llama-3.2-3B-Instruct} & \textbf{Qwen3-4B} & \textbf{Qwen3-8B} \\ \midrule
\multirow{2}{*}{\textbf{MedQA}}      & KVC ~$\uparrow$              & 0.418                          & 0.320             & 0.001             \\
                                     & KITE ~$\downarrow$             & -0.616                         & -0.860            & -0.910            \\ \midrule
\multirow{2}{*}{\textbf{ARC-C}}      & KVC ~$\uparrow$              & 0.242                          & 0.231             & 0.227             \\
                                     & KITE ~$\downarrow$             & -0.547                         & -0.999            & -0.884            \\ \midrule
\multirow{2}{*}{\textbf{HumanEval+}} & KVC ~$\uparrow$              & 0.423                          & 0.482             & 0.079             \\
                                     & KITE ~$\downarrow$             & -0.620                         & -0.691            & -0.837            \\ \midrule
\multirow{2}{*}{\textbf{AIME24}}     & KVC ~$\uparrow$              & 0.700                          & 0.477             & -0.004             \\
                                     & KITE ~$\downarrow$             & -0.813                         & -0.570            & -0.789            \\ \midrule
\multirow{2}{*}{\textbf{Average}}    & KVC ~$\uparrow$              & 0.445                          & 0.377             & 0.075             \\
                                     & KITE ~$\downarrow$             & -0.649                         & -0.780            & -0.855            \\ \bottomrule[1.2pt]
\end{tabular}%
}
\caption{Pearson correlation coefficients between layer selection scores and accuracy under single-layer KV communication. For each model, we randomly sample 10 layers. KVC computes attention-based scores independently for each task. KITE computes trajectory-distortion scores on the calibration set and reuses them across all tasks. For KITE, MedQA is in-domain, while ARC-C, HumanEval+, and AIME24 are out-of-domain.}
\label{tab:correlation}
\end{table}

\begin{figure}[h]
\centering
\includegraphics[width=0.95\columnwidth]{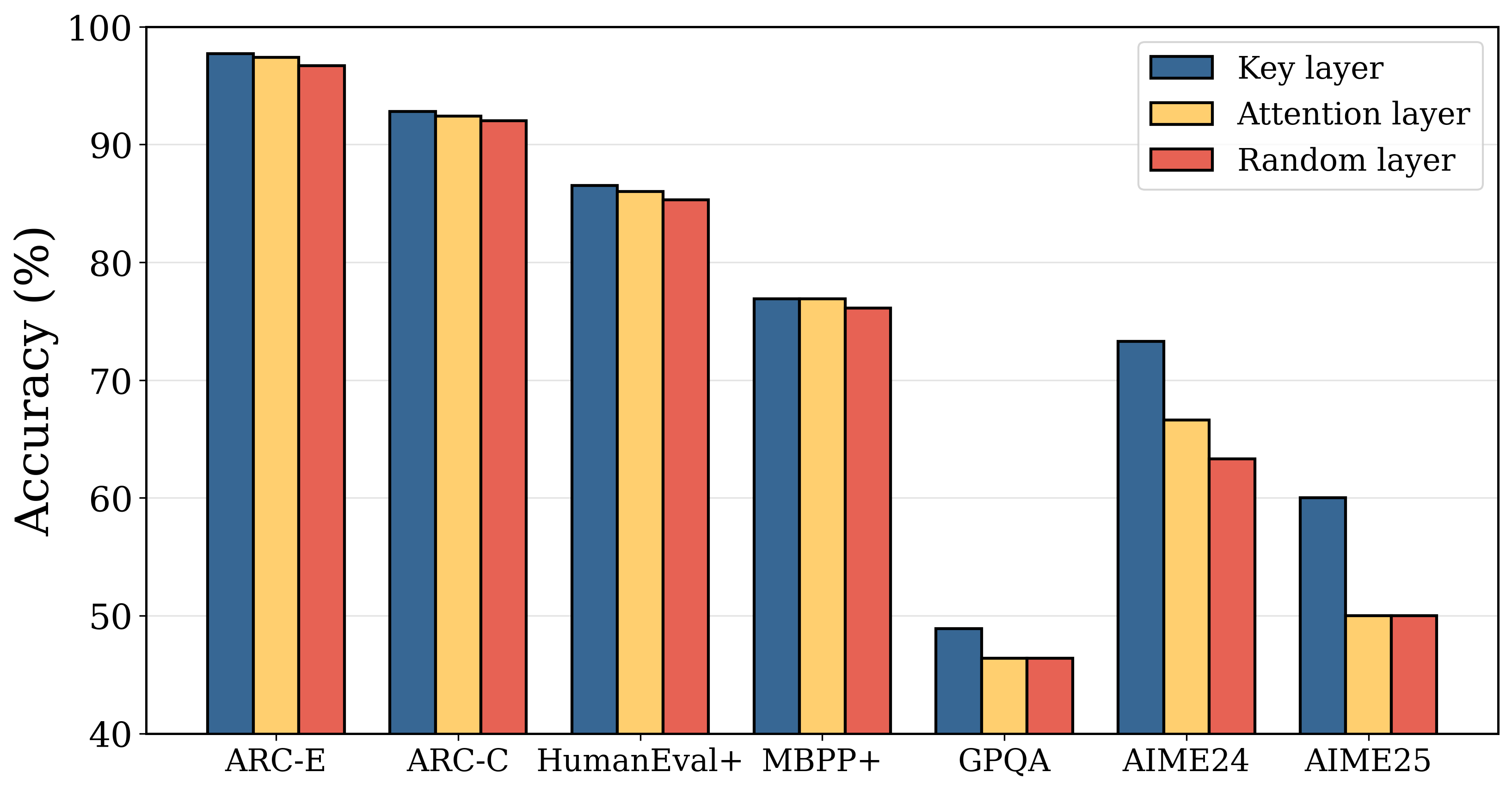} 
\caption{Comparison among key-layer, attention-selected, and random-layer communication using Qwen3-4B. The attention layer denotes the highest-scoring KV layer selected independently for each task using attention-based criterion in KVC.}
\label{fig:random_layer_vs_key_layer}
\end{figure}

We further validate the task effectiveness of the identified key layer. We compare the key layer with both a randomly selected layer and the highest-scoring layer selected by KVC. As shown in Figure \ref{fig:random_layer_vs_key_layer}, key-layer communication consistently outperforms attention-selected and random-layer communication across all tasks. The performance gaps are relatively small on easier tasks, where limited reliance on inter-agent collaboration reduces sensitivity to layer selection. In contrast, the key layer yields larger gains on more challenging tasks, demonstrating that our criterion identifies layers carrying task-effective  information for communication.

\label{sec:Selection_Strategy_Effectiveness}

\subsection{Optimal Number of Communicated KV Layers}
\label{sec:layer_num}
\begin{figure}[h]
\centering
\includegraphics[width=0.95\columnwidth]{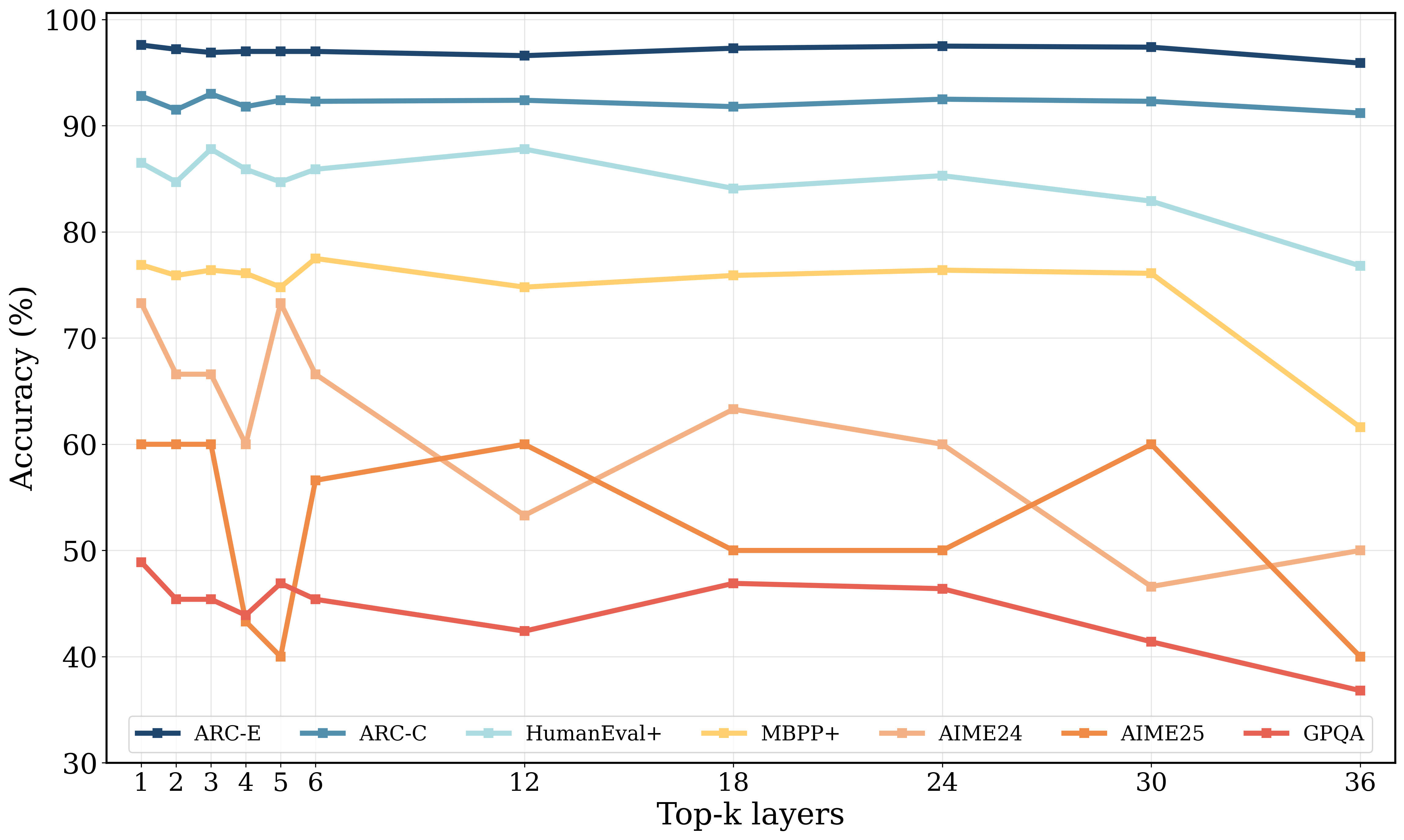} 
\caption{Accuracy of Qwen3-4B under top-$k$ KV layers communication.}
\label{fig:top-k_accuracy}
\end{figure}

To understand how many KV layers are required for optimal performance in latent communication, we vary the number of communicated KV layers, selecting the top-k layers according to the layer ranking and evaluating them on seven downstream tasks. As shown in Figure \ref{fig:top-k_accuracy}, increasing the number of communicated layers generally degrades task performance, indicating that not all layers’ KV cache are task-effective. Since communicating only the top ranked key layer achieves the best overall accuracy, we set the selected subset size to 1 in KITE, jointly optimizing task performance and communication efficiency.

\subsection{Effect of Latent Injection Position}

\begin{figure}[h]
\centering
\includegraphics[width=0.95\columnwidth]{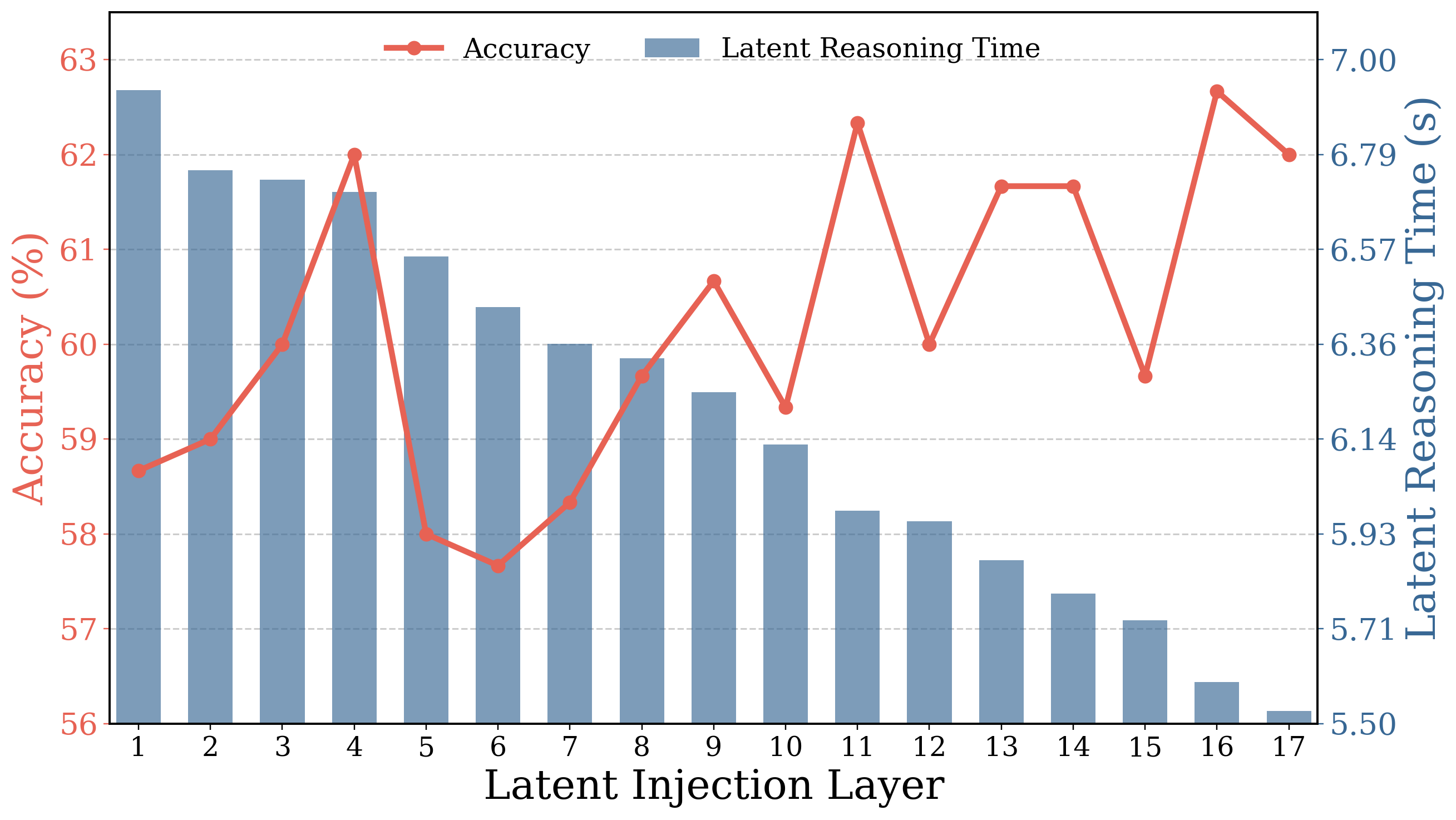} 
\caption{Effectiveness of different latent injection layers with Qwen3-4B on MedQA calibration set.}
\label{fig:latent_injection-layer_ablation}
\end{figure}

We fix the communicated key layer and inject the last hidden states into different layers for autoregressive latent reasoning, in order to analyze the effect of the latent injection position. Figure \ref{fig:latent_injection-layer_ablation} reports both accuracy and the total latent reasoning time of the multi-agent system. As the injection layer moves deeper, latent reasoning time decreases steadily, since fewer layers are computed at each latent reasoning step. In contrast, accuracy fluctuates across injection layers. Injecting into middle layers generally yields higher accuracy than injecting into shallow layers. Notably, injecting at the key layer 17 achieves competitive accuracy while incurring the lowest reasoning time, indicating that aligning the latent reasoning entry point with the key communication layer provides a favorable balance between task performance and computation efficiency.

\subsection{Analysis of the Performance Gains}
\label{accuracy improvement analysis}
\begin{figure}[h]
\centering
\includegraphics[width=0.95\columnwidth]{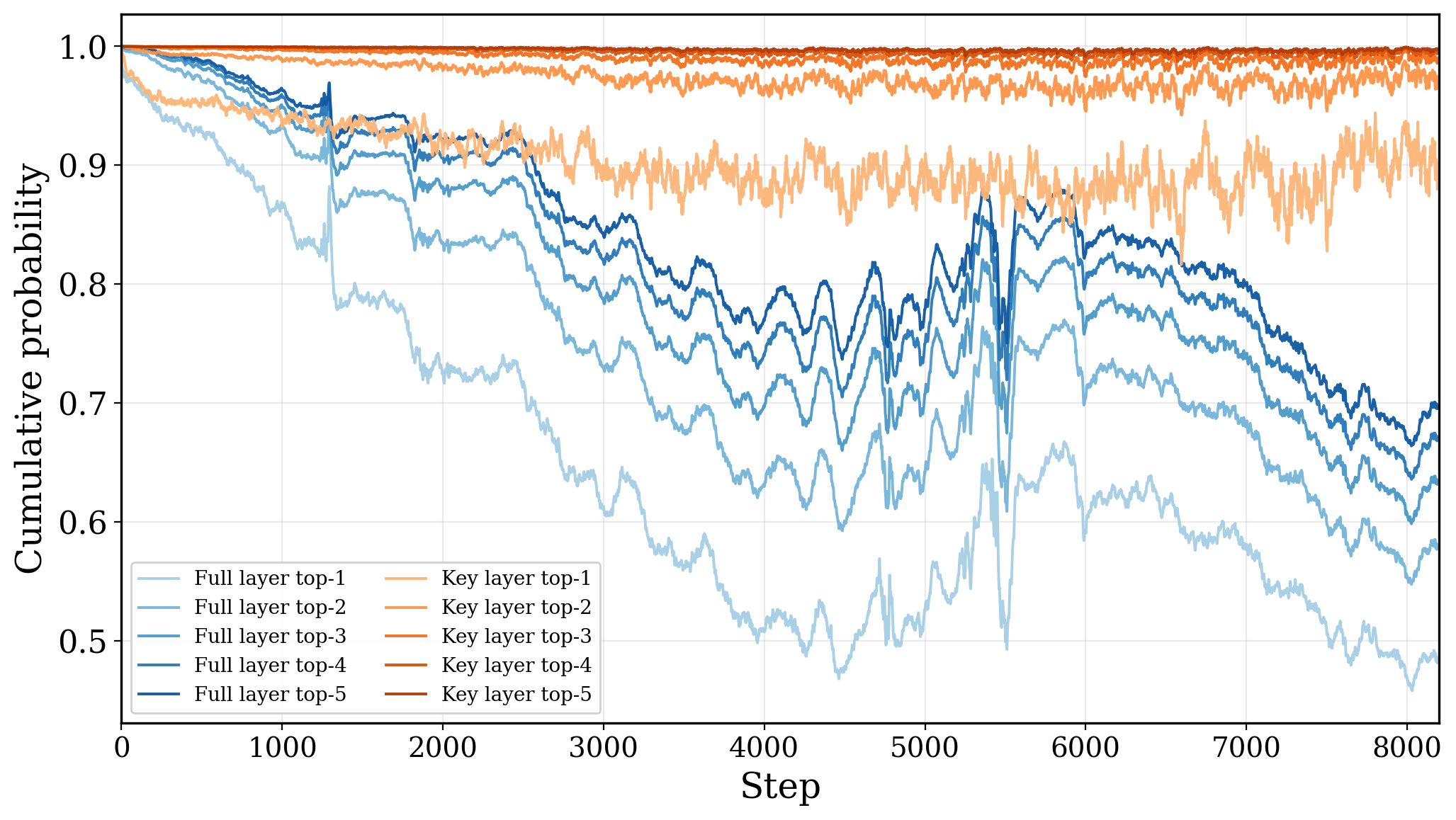} 
\caption{Cumulative probability mass of the top-$k$ tokens during solver agent $A_4$ decoding on HumanEval+ with Qwen3-4B. We compare full-layer KV communication with key-layer communication for $k=1, 2, 3, 4,5$.}
\label{fig:top-5-probabilities}
\end{figure}

We analyze the performance gains of key-layer communication through the solver’s token-level decoding confidence. As shown in Figure \ref{fig:top-5-probabilities}, KITE consistently maintains substantially higher cumulative probability mass: the top-1 token probability remains around 90\%, while the cumulative top-5 probability approaches 1 throughout decoding. By contrast, under full-layer communication, the probability mass gradually decreases and exhibits larger fluctuations, with the top-1 probability falling below 50\% and the cumulative top-5 probability dropping below 70\%. This more dispersed output distribution shows greater uncertainty in the solver’s token predictions. These results suggest that communicating only the key-layer KV cache may act as an implicit information bottleneck, preserving task-effective  signals while suppressing redundant or conflicting information encoded at other layers. This may enable the receiver to maintain a more coherent reasoning trajectory and make more reliable decisions toward task completion. Together, these observations indicate that task-oriented latent communication can improve not only efficiency but also downstream performance by filtering unnecessary internal states.





\section{Conclusion}
In this work, we revisit latent multi-agent communication from a task-oriented perspective and propose KITE, a training-free framework for efficient LLM-based multi-agent collaboration. KITE identifies a task-effective key layer through receiver trajectory distortion, communicates only the latent working memory associated with this layer, and uses the same layer as the entry point for latent reasoning. Extensive experiments on diverse benchmarks and models demonstrate that KITE reduces communication volume by 28\text{--}36$\times$, accelerates end-to-end inference by up to 3$\times$, and improves accuracy by up to 23.3 percentage points over full-layer KV communication. These results highlight task-oriented communication as a promising direction for efficient multi-agent collaboration.

\bibliography{aaai2027}

\clearpage
\appendix

\section{Experiment Setups}

we provide more details about the experimental setup, including dataset details and implementation details.
\subsection{Dataset}

\textbf{Math \& Science Reasoning.}

\textbf{MedQA} \cite{medqa} is a medical question-answering benchmark derived from professional medical licensing examinations. It evaluates biomedical knowledge, clinical reasoning, and diagnostic decision-making through multiple-choice questions that require models to integrate domain-specific knowledge with the information provided in each clinical scenario.

\textbf{GSM8K} \cite{gsm8k} is a widely used benchmark containing approximately 8.5K grade-school mathematical word problems. Each problem typically requires multiple arithmetic reasoning steps to transform a natural-language description into a numerical answer, making it a standard testbed for assessing chain-of-thought reasoning ability.

\textbf{GPQA-Diamond} \cite{gpqa} is the most challenging split of the Graduate-Level Google-Proof Q\&A benchmark. It contains 198 expert-written multiple-choice questions in physics, biology, and chemistry. The questions are designed to be difficult to answer through simple retrieval and instead require graduate-level domain knowledge, conceptual understanding, and multi-step scientific reasoning.

\textbf{AIME24} \cite{maxwelljia2024aime} contains 30 problems from the 2024 American Invitational Mathematics Examination. The problems cover algebra, geometry, number theory, and combinatorics. Due to its competition-level difficulty and carefully curated problem set, AIME24 provides a challenging evaluation of precise multi-step mathematical reasoning.

\textbf{AIME25} \cite{mathai2025aime} consists of 30 problems from the 2025 American Invitational Mathematics Examination. It provides an additional set of Olympiad-style problems involving algebraic manipulation, geometric reasoning, number theory, and combinatorial analysis.

\noindent \textbf{Commonsense Reasoning.}

\textbf{ARC-Easy} and \textbf{ARC-Challenge} \cite{arc} are two subsets of the AI2 Reasoning Challenge, a collection of multiple-choice science questions drawn from standardized school-level examinations. ARC-Easy contains questions that are generally answerable through basic scientific knowledge and relatively direct reasoning, whereas ARC-Challenge consists of more difficult questions that typically require multi-step reasoning, knowledge integration, and discrimination among plausible distractors.

\noindent \textbf{Code Generation.}

\textbf{MBPP-Plus} \cite{mbppANDhumaneval} is an enhanced version of the MBPP benchmark. It contains short natural-language programming tasks paired with function signatures and test cases, while the MBPP-Plus augments the original benchmark with more extensive and rigorous tests.

\textbf{HumanEval-Plus} \cite{mbppANDhumaneval} extends the original HumanEval benchmark with substantially strengthened test cases for each programming problem. Each instance provides a natural-language function specification and requires the model to generate a correct Python implementation. The additional tests expose solutions that pass the original HumanEval tests but fail under more comprehensive functional evaluation, making HumanEval-Plus a stricter benchmark for code correctness and generalization.

\subsection{Implementation Details}

We use the official chat templates and special tokens provided with each model. For Qwen models, we adopt the corresponding special tokens, including \texttt{<|im\_start|>} and \texttt{<|im\_end|>}, while for Llama models, we use \texttt{<|start\_header\_id|>} and \texttt{<|end\_header\_id|>}. All experiments are conducted on an NVIDIA RTX 5880 Ada Generation GPU with 48 GB of memory.

\paragraph{Evaluation Protocol.}
We report accuracy for all non-coding benchmarks based on answer matching after normalizing the final response by lowercasing, trimming whitespace, and removing extraneous punctuation. For multiple-choice datasets, including GPQA-Diamond, MedQA, ARC-Easy, and ARC-Challenge, we extract the final predicted answer and compare it with the ground-truth option letter using exact matching. For numerical reasoning datasets, including GSM8K, AIME24, and AIME25, we extract the final predicted value, parse both the prediction and ground-truth answer as numbers, and determine correctness by numerical equality. Predictions that cannot be parsed as valid numbers are counted as incorrect.

For code-generation benchmarks, including HumanEval-Plus and MBPP-Plus, we evaluate functional correctness by executing the generated code against the benchmark test cases. Specifically, we extract the predicted code from the model output, combine it with the provided unit tests, and execute the resulting program in a sandboxed environment with a 10-second timeout. A sample is considered correct only when all test cases pass without runtime errors.

\section{Positional Embedding Coherence}
We maintain a consistent positional frame across all layers when communicating only the key-layer KV cache. Let \(c\) denote the length of the incoming latent working memory and \(n\) the receiver's prefill length. We shift the receiver's position indices by \(c\) at every layer. At the selected key layer \(l^*\), the sender's KV cache occupies positions \([0,c)\), while the receiver's KV cache is appended at positions \([c,c+n)\). At non-key layers, positions \([0,c)\) contain no KV states, but the receiver tokens still start from position \(c\). Consequently, the receiver uses the same position indices at every layer, avoiding positional misalignment across the model depth. During subsequent autoregressive generation, including latent reasoning and final token decoding, newly generated latent states continue from position \(c+n\).

\section{Additional Analyses on Selection Strategy}

\subsection{Effect of Calibration Set Size}

\begin{figure}[h]
\centering
\includegraphics[width=0.95\columnwidth]{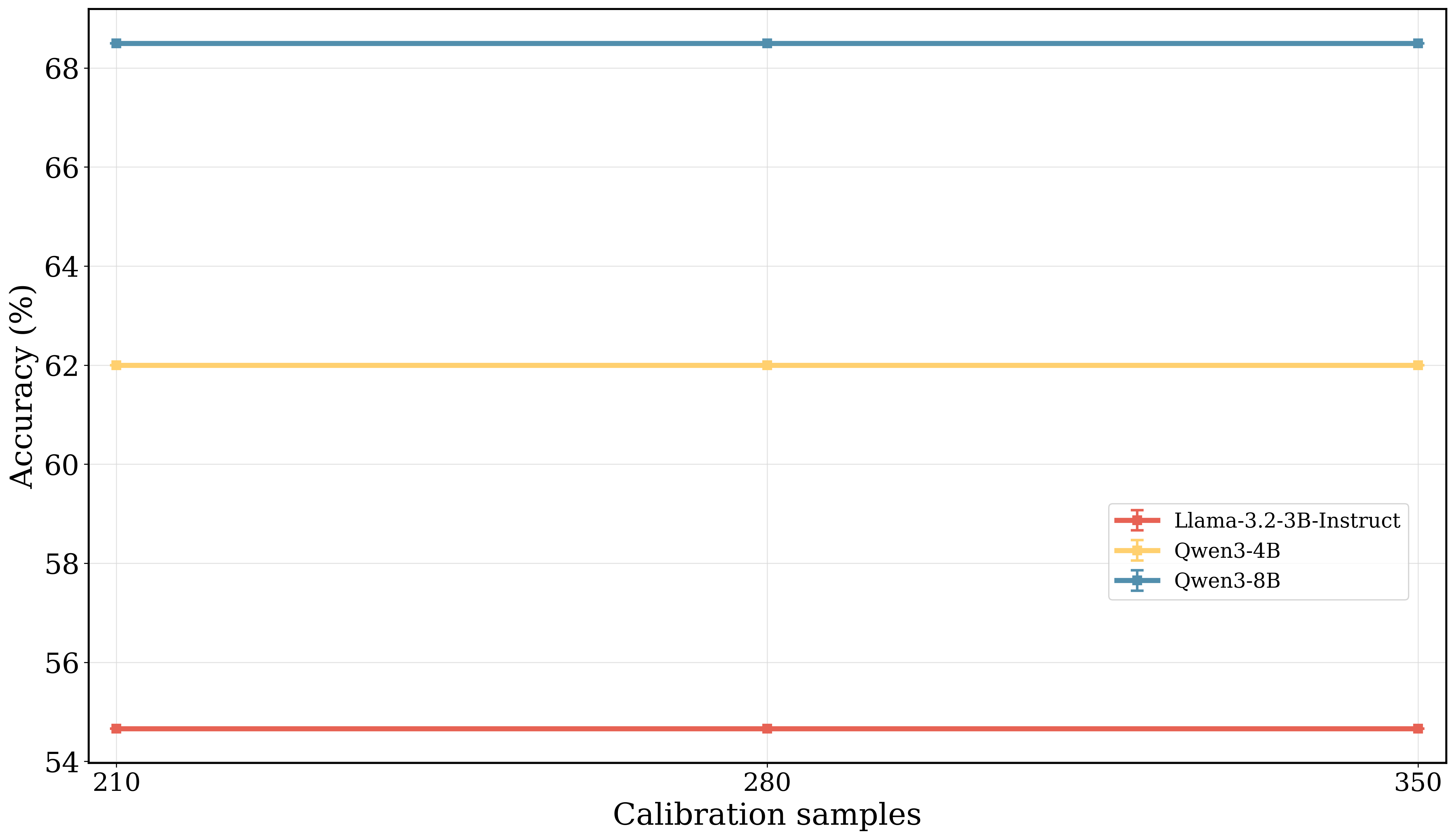} 
\caption{Effect of calibration set size on key-layer selection.}
\label{fig:selection_samples}
\end{figure}

To investigate the effect of calibration set size on key-layer selection, we vary the number of calibration samples and perform proportional stratified sampling from the full calibration set. For each calibration size, we repeat the random sampling three times, rank all layers using the sampled data, and evaluate the highest-ranked layer on the MedQA calibration set. As the results in Figure~\ref{fig:selection_samples} show, calibration set size has little effect on downstream performance: all evaluated sample sizes achieve the same accuracy across the three backbone models. These results suggest that the proposed selection strategy is robust to the calibration set size.

\subsection{Effect of Trajectory Distortion Metrics}

\begin{table}[h]
\centering
\resizebox{0.48\textwidth}{!}{%
\begin{tabular}{lcccc}
\toprule[1.2pt]
\textbf{Tasks}                       & \textbf{Metrics} & \textbf{Llama-3.2-3B-Instruct} & \textbf{Qwen3-4B} & \textbf{Qwen3-8B} \\ \midrule
\multirow{3}{*}{\textbf{MedQA}}      & KL ~$\downarrow$              & -0.501                          & -0.864             & 0.694             \\
& Similarity ~$\uparrow$              & 0.609                          & 0.859             & 0.847             \\
                                     & MSE ~$\downarrow$             & -0.616                         & -0.860            & -0.910            \\ \midrule
\multirow{3}{*}{\textbf{ARC-C}}      & KL ~$\downarrow$             & -0.480                          & -0.996             & 0.728             \\
& Similarity ~$\uparrow$              & 0.543                          & 0.999             & 0.854            \\
                                     & KITE ~$\downarrow$             & -0.547                         & -0.999            & -0.884            \\ \midrule
\multirow{3}{*}{\textbf{HumanEval+}} & KL ~$\downarrow$             & -0.328                          & -0.693             & 0.722             \\
& Similarity ~$\uparrow$              & 0.620                          & 0.691             & 0.820             \\
                                     & KITE ~$\downarrow$             & -0.620                         & -0.691            & -0.837            \\ \midrule
\multirow{3}{*}{\textbf{AIME24}}    & KL ~$\downarrow$              & -0.573                          & -0.577             & 0.758             \\
& Similarity ~$\uparrow$              & 0.817                          & 0.570             & 0.836             \\
                                     & KITE ~$\downarrow$             & -0.813                         & -0.570            & -0.789            \\ \midrule
\multirow{3}{*}{\textbf{Average}}  &  KL ~$\downarrow$             & -0.470                          & -0.782             & 0.725             \\
& Similarity ~$\uparrow$              & 0.647                          & 0.779             & 0.839             \\
                                     & KITE ~$\downarrow$             & -0.649                         & -0.780            & -0.855            \\ \bottomrule[1.2pt]
\end{tabular}%
}
\caption{Pearson correlation coefficients between layer-ranking scores and accuracy under single-layer KV communication using different trajectory distortion metrics. We compare KL divergence (\textbf{KL}), cosine similarity (\textbf{Similarity}), and the MSE (\textbf{MSE}).}
\label{tab:metrics_correlation}
\end{table}

To examine the effect of different trajectory distortion metrics, we compare KL divergence, cosine similarity, and MSE by measuring the correlation between their layer-ranking scores and task accuracy under single-layer KV communication. As shown in Table~\ref{tab:metrics_correlation}, the MSE-based criterion used in KITE consistently exhibits strong correlations in the expected direction across all model--task combinations. In contrast, KL divergence produces positive correlations on Qwen3-8B, which conflicts with its lower-is-better interpretation and indicates less stable layer ranking. Cosine similarity also achieves strong correlations, but MSE provides better overall performance, obtaining the strongest average correlations on the three models. These results indicate that MSE is a stable and effective metric for measuring receiver trajectory distortion and identifying task-effective KV layers.

\section{Prompts}

\begin{figure}[h]
\centering
\includegraphics[width=0.48\textwidth,trim=0 500 0 0,clip]{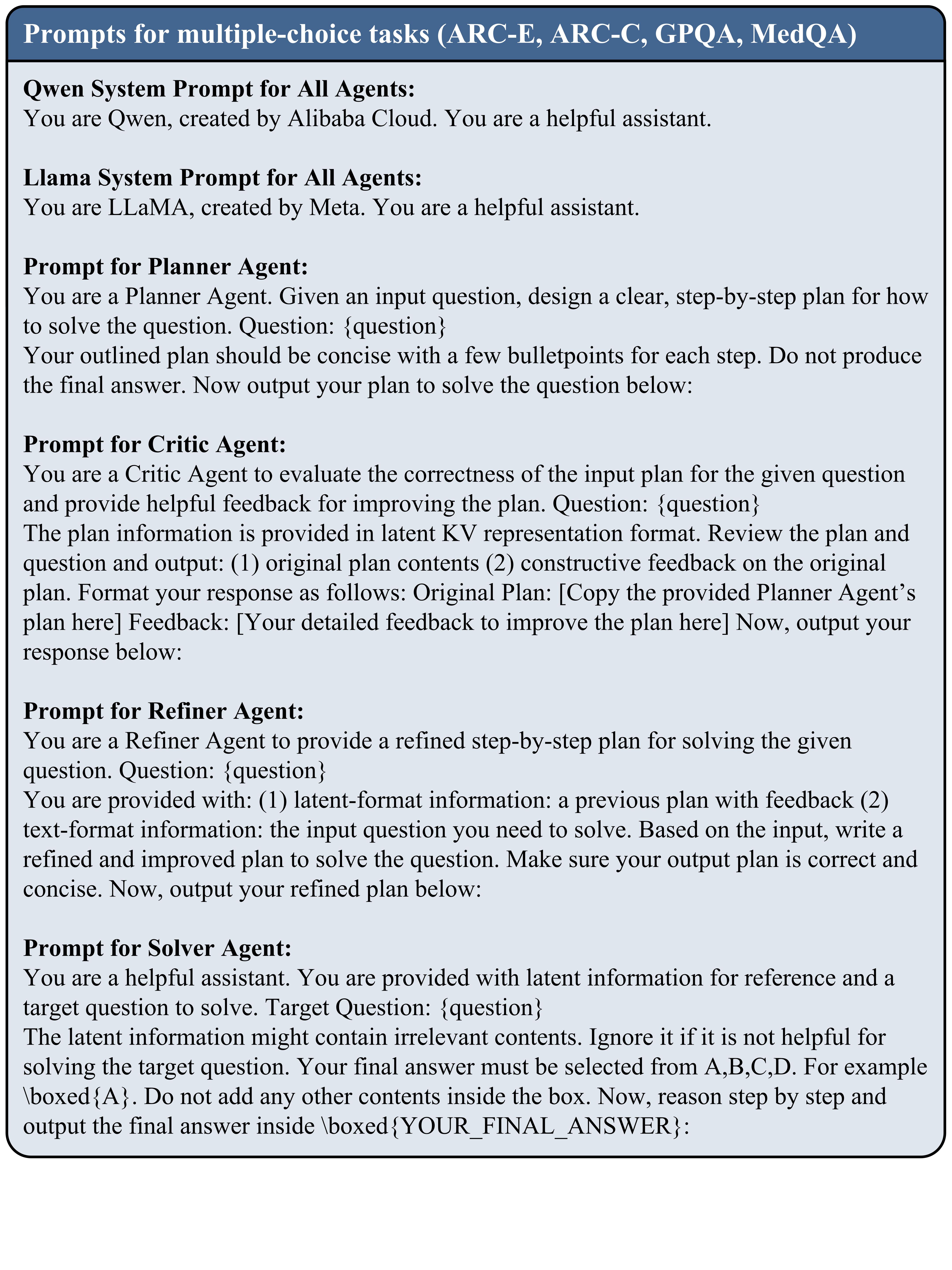} 
\caption{Prompts for multiple-choice tasks.}
\label{fig:prompt1}
\end{figure}

The planner, critic, and refiner use the same role-specific prompts across all tasks, as shown in Figure~\ref{fig:prompt1}. The solver prompt is adapted to the required output format of each task.

\begin{figure}[H]
\centering
\includegraphics[width=0.48\textwidth,trim=0 2300 0 0,clip]{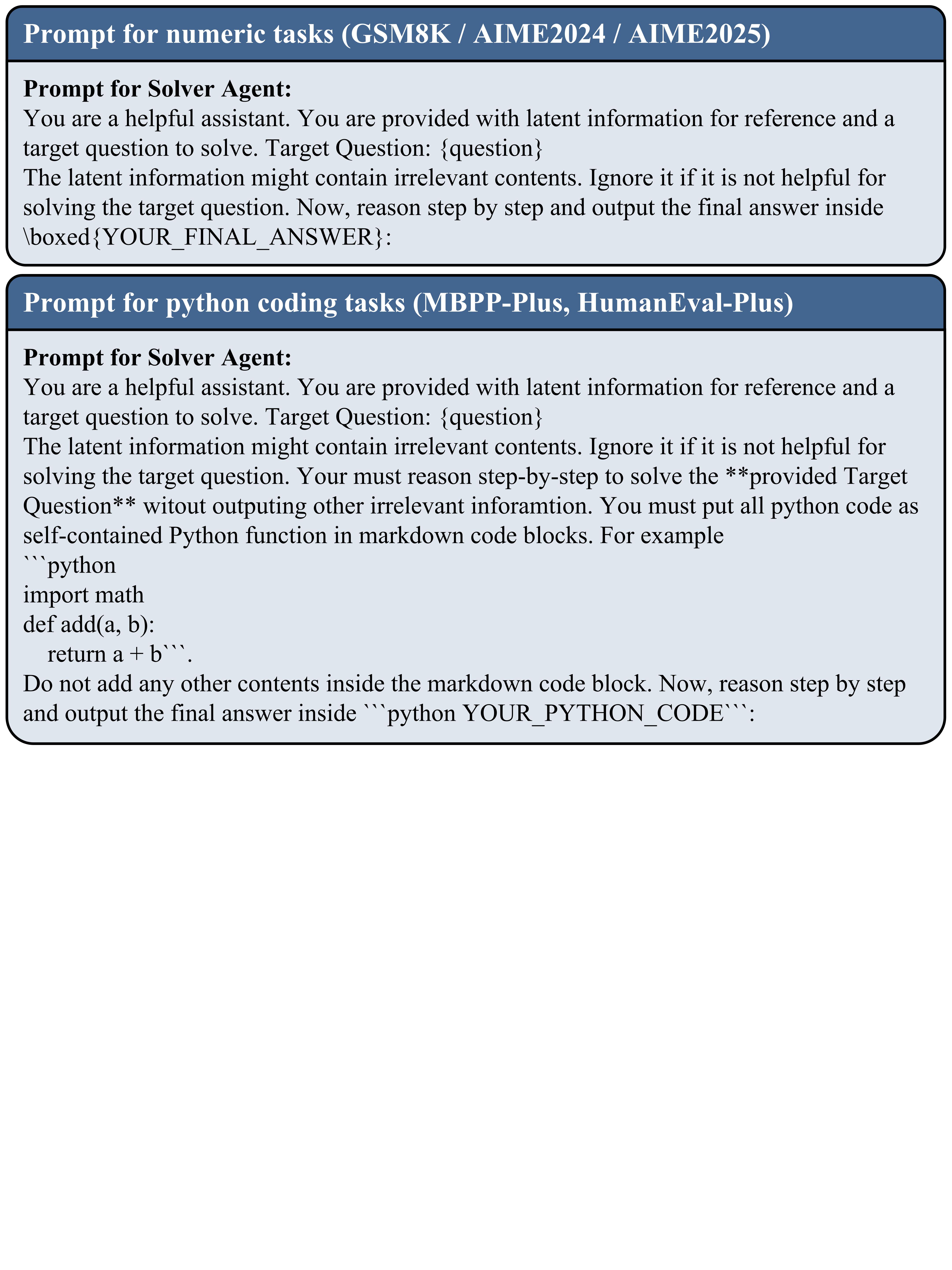}
\caption{Solver prompts for numeric and cooding tasks.}
\end{figure}

\end{document}